# 3 Towards Open Standards for Systemic Complexity in Digital Forensics

*Paola Di Maio*

## 3.1 UBIQUITY IN DIGITAL FORENSICS

Digital forensics (DF), in the context of forensic science, is concerned with identifying, gathering, storing, analyzing, and communicating digital evidence from electronic devices and computer systems [1]. Its core capability is based on precise information acquisition techniques and logical processing functions, in particular, deductive reasoning.

Sophisticated technologies are now used to support fact-checking and inquiry, not only pertaining to criminal investigations but also in scientific and technical inquiry. Given that in our times all processes are mediated by digital technologies, DF techniques can be used to reconstruct and understand facts and events generated in digital scenarios, as well as in all cases where digital records of non-digital (physical) evidence are used to support legal, scientific, or technical analyses of all kinds.

As such, DF is increasingly relevant to all fields of inquiry and ubiquitous. On the one hand, it consists of processes, techniques, and algorithms applicable to digital technologies supporting the analysis of both digital and non-digital (digital records of physical) evidence, and on the other hand, it consists of machine learning techniques that automate these logical analysis and inferencing processes (Figure 3.1).

As technology evolves, and DF becomes exponentially complex, its principles and techniques become applicable to diverse areas of practice, such as in the responsible management of digital records, digital policy compliance, and reproducible data science. The convergence of artificial intelligence (AI) and DF provides powerful data analysis capabilities, especially the facility to automate the screening of large datasets. Nonetheless, the use of AI techniques in forensics comes with its own challenges and risks that can only be addressed if the underlying logical processes are standardized, transparent, explainable, and reproducible. It is impossible to evaluate the correctness of DF if the data and processes are not standardized or transparent.

This chapter contributes central arguments to the development of open standards in DF to analyze, correlate, and query data stores using explainable AI with the intent to ascertain truth, not only for the purpose of law enforcement but also for organizational governance and scientific inquiry in general, pointing to the need for greater transparency and accountability supported by the human readability of digital and machine learning artifacts.

                                                                33

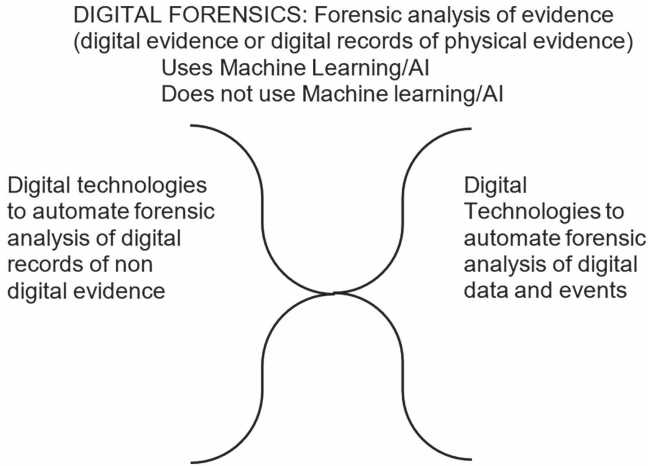

**FIGURE 3.1** Ubiquity of Digital Forensics.

Despite the usefulness of novel AI tools, it is ultimately human experts and adjudicators who validate and interpret the outcomes of automated analysis, a practice that is sometimes referred to as humans in the AI loop [2]. The ability of humans to assess the validity of automated processes requires human-readable digital artifacts. To mitigate bias and demonstrate accountability, and as such, the validity of the outcomes of automation, both the data and the processes need to come out of the black box. The approach proposed in this chapter advocates for open human-readable standards in DF.

## 3.2 SYSTEMIC COMPLEXITY

Thanks to the proliferation of technologies for data storage and analysis, and to widening fields of application, digital forensics (DF) is expanding and now overlaps with other disciplines, leading to a state of **systemic complexity**, defined in terms of its multidimensional evolution and rapid reconfiguration of its functional, systemic, and behavioral boundaries. Technologies based on complex knowledge domains can be powerful yet fragile at the same time. This research identifies this complexity and addresses it with a system-level knowledge representation approach.

It can be said that whatever DF literature typically identifies as forensic evidence can actually be reduced to what computer science refers to as **data**, resulting in an inevitable overlap of DF with CS and Data Science (DS) (Figure 3.2).

Forensic tools and techniques have become necessary to sustain robust inquiry not only in law enforcement but also in every kind of scientific investigation, especially in data science (DS). Due to the exponential proliferation of datasets from research, their size, and heterogeneity as found in typical analytical scenarios, data curation practices require accurate and systematic techniques for the collection, maintenance, and analysis of data, which are instrumental in ensuring the integrity, accessibility, and validity of any finding. For example, data management planning (DMP)[1] methods prescribe how to systematically collect, store, secure, archive, and share data, with their adoption having become a standard requirement in project management and scientific research.

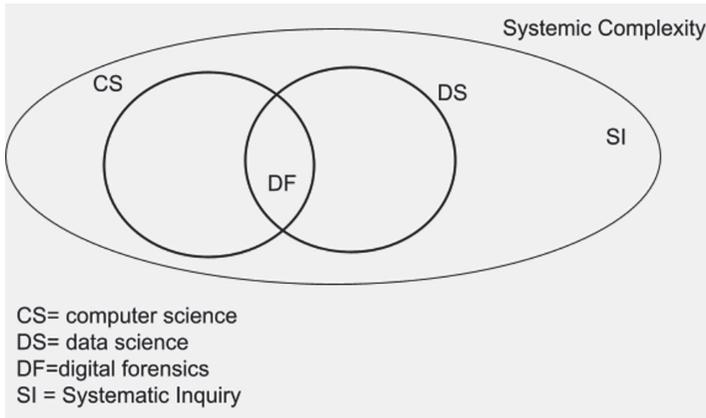

**FIGURE 3.2**  Systemic Complexity in Digital Forensics.

The evaluation of data quality and integrity is central to assessing the accuracy of DF, which also depends on the adoption of well-defined and commonly used data formats for interoperability [3]. Data extraction, pattern recognition, and anomaly detection are DF techniques commonly used in DF and in data science [4].

## 3.3  DIGITAL FORENSICS AND ARTIFICIAL INTELLIGENCE

Through DF, electronic data can be analyzed in detail to determine the chronological and factual accuracy of information lifecycles, thus enabling their interpretation in support of hypotheses, a logical process of evaluation and analysis that can be enhanced by the use of artificial intelligence (AI).

Early efforts to record and examine individual incidents of user behaviors were ad-hoc and unsystematic; however, together with technological advancement, DF progressively developed towards standardized practices, leading to the development of foundational concepts and to the emergence of a digital forensics community working towards legal frameworks [5].

The continuous evolution of digital forensics aims to achieve stable standardized methodologies and best practices, with emerging specializations from mobile forensics [6] to network forensics [7].

Nowadays, DF advanced technologies are being used for data recovery, analysis, and extraction, and to expose registry information. AI-supported DF leverages machine learning to automate and optimize specific forensic techniques, such as data carving, to reconstruct deleted or fragmented data from storage devices, steganography, stochastic analysis, and file deletion.

Once the data has been identified, cleaned up, and verified, DF tools can be designed to support its use to perform intelligent reasoning and reach conclusions.

Likewise, any misinterpretation and erroneous reconstruction of fragmented or missing datasets can lead to spurious fabrication, justified and supported by the latest technology.

By its programmatic nature, computational environments offer native support to certain analyses, thanks to their systematic processes and protocols, and their inherent logging capabilities. Typical built-in logging utilities that can be leveraged by DF include **system logs**, used to track system start-up, shutdown, hardware events, and software installations; **application logs** that record application activity, errors, and user interactions; **security logs** that capture login attempts, access control events, and security incidents; and **network logs** tracking network traffic, including connection attempts, data transfer, and firewall activity.

Other analytical artifacts that can support forensic analysis include System Resource Usage Monitor (SRUM), Prefetch, and, in Windows OS, ShimCache and *AmCache*, which are *Windows artifacts* that *contain* information about recently executed *applications*.

AI and ML techniques can be particularly advantageous in parsing, analyzing, and correlating data from these diverse sources.

DF architectures can be designed based on logical and functional layers to enable the systematic collection and analysis of data chain of custody to validate assertions of statements of facts in diverse areas of inquiry such as incident response, electronic discovery, corporate investigations, data recovery, and more generally to support evidence-based rational inquiry. AI technologies can be especially useful to facilitate big data analysis [8, 9].

The integration of artificial intelligence (AI) into digital forensics delivers enhanced efficiency in terms of depth and breadth of analysis, as well as speed. AI-powered tools can automate and accelerate exponentially repetitive tasks like data acquisition, processing, and analysis [10], while saving time and resources. By the same token, the rapid automation of analytical processes can enhance the fallacies, flaws, and errors in analytical models and outcomes, exploding the potential rate of error. To identify and correct AI-generated errors, it is necessary to employ highly skilled human experts through manual or semi-manual verification, which can be very costly or even unfeasible at a large scale.

Machine learning algorithms, in principle, can improve the accuracy of results by sifting through vast amounts of data to identify patterns and anomalies, analyze complex relationships between disparate data points that might be missed by human analysts, leading to more accurate findings. Typical applications include the automation of malware and anomaly detection, the analysis of code features and network behavior to identify malicious activities, and predictive analytics, which can forecast potential breaches based on historical data and user behavior patterns [11].

Among the latest ML techniques in use, *memetic algorithms* combine evolutionary algorithms with local solvers referred to as memes [12], and *fronesis* consists of rule-based reasoning and attack detection rules to identify adversarial techniques [13], in addition to deep learning techniques such as "stacked auto-encoders (SAE), deep belief networks (DBN), convolutional neural networks (CNN), and recurrent neural networks (RNN)" [14].

Despite the proliferation of ML techniques and their growing adoption, strong AI is not yet part of the DF process due to innumerable challenges, discussed in the next section, which can contribute to incorrect outcomes, wrong conclusions, suboptimal or wrong decisions, and misjudgment. The challenges to data and process integrity exist in all forensic analyses but can be magnified exponentially when using machine learning.

Research identifies possible approaches to resolve the challenges [15], for example, recommending the use of explicit knowledge representation techniques such as pattern recognition and expert systems to enhance ML capabilities and processes and "ensure that their deployment is ethical, transparent, and respectful of data subjects' rights and privacy" [11].

The chapter contributes directly towards such recommendations.

## 3.4  OPEN CHALLENGES

The overall challenge and limitation of forensic sciences is the ability to consider appropriate evidence and follow a logical process analysis to reach correct conclusions to support appropriate judgments and decisions. Although DF brings advantages in terms of scale and processing capacity, the use of ML can also weaken the scrutability of processes because of a lack of transparency. A substantial body of knowledge documents how forensic science gone wrong contributes directly to miscarriage of justice (MoJ) [16].

Experts have been calling for "uncompromising mechanisms capable of identifying flaws in the system, instigating appropriate remedial action in particular cases, and ensuring that necessary reforms are implemented to mitigate misjudgement" [17]. The adoption of DF aims to establish verifiable truth of data and processes that ascertain facts and the legitimate adherence to protocols.

Nonetheless, the final validation of the outcomes, to assess their accuracy, integrity, and accountability, still depends entirely on human judgment. DF practitioners typically apply categorical conclusions or strength of support conclusion types that cannot really be verified. Both humans and AI, however, are affected by bias [18]. Forensic science and its practitioners are respected and valued in judicial proceedings due to their capacity to discover irrefutable facts that may otherwise not come to light. But they are not immune from logical fallacies (Figure 3.3) [19].

In particular, confirmation bias is known as "a proclivity to search for or interpret additional information to confirm beliefs and to steer clear of information that may disagree with those prior beliefs" [20]. Blind trust in the correctness of software-generated evidence is known to have resulted in wrongful convictions for decades [21]. There are expectations that AI may help reduce human bias by standardizing the reasoning; however, machine learning-generated bias is a concern [22]. Although scientific methods can be used to assert and evaluate different degrees of confidence and even certainty in the outcome of automated processes, machine learning itself tends to lack an explanation of its meaning or reference to established frameworks.

"Substantial deficiencies in documentation practices for content essential for enabling audit of the DF investigative process and results, a challenge shared with other forensic science disciplines" [23].

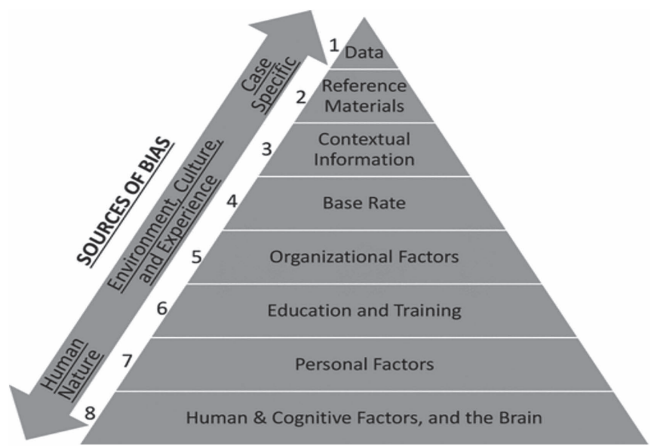

**FIGURE 3.3**   Six Fallacies and Eight Sources of Bias Dror, I. E. (2020).

The effectiveness of ML models in DF depends on the quality and representativeness of training data. Biases embedded in data can skew results and result in unreliable conclusions. The ultimate logical composition and analysis of the evidence/data identified through the DF process remain a prerogative of the intelligent human expert, capable of establishing the relevance of the evidence, interpreting, and drawing conclusions from the analysis. Understanding and addressing both human and machine bias is non-trivial.

Bias is at the root of all human and machine intellectual and cognitive limitations, and there are countless ways that it can be characterized, starting from philosophical disquisition to mathematical, geometric, and statistical applications [24]. Understanding the reasoning behind AI-generated evidence and decisions remains a challenge, potentially hindering trust and legal acceptance [15].

Integrating AI into forensics raises additional concerns about data security, privacy breaches, and potential misuse of powerful algorithms [25]. Researchers warn that the legal admissibility of AI-generated evidence and the ethical implications of using AI in investigations require careful consideration [26].

There is general acceptance that AI technologies are increasing the efficiency of analyses; however, their reliability is limited due to a lack of transparency [27]. The landscape of challenges for DF is vast and complex. It can be traversed using taxonomic approaches, for example, by identifying categories such as technical, legal, personnel, and operational [25].

**USE OF TAXONOMIES IN DF RESEARCH (from Karie Venter)**

- Altschaffel et al.: Digital forensic investigations are usually conducted to solve crimes committed by perpetrators and/or intruders using IT systems. They then propose a taxonomy that helps perform a forensic examination and to establish answers to a set of well-defined questions during such examination.

- Hoefer and Karagiannis: Taxonomy of cloud computing services. A tree-structured taxonomy based on their characteristics, to easily classify cloud computing services and make it easier to compare them. In contrast, the proposed taxonomy in this paper offers a simplified platform that sheds more light on the classification of existing digital forensic challenges.
- Strauch et al.: Taxonomy to categorize existing cloud data-hosting solutions.
- Lupiana et al.: Taxonomy for classifying disparate research efforts in ubiquitous computing environments. Their taxonomy classifies ubiquitous computing environments into two major categories: interactive environments and smart environments.
- Sansurooah: Methodologies and procedures that are in place for the gathering and acquisition of digital evidence and subsequently defines which model will be the most appropriate taxonomy for the electronic evidence in the computer forensics analysis phase.
- Sriram: Reviews the research literature since 2000 and categorizes developments in the field into four major categories.
- Kara et al.: New research categories (taxonomy) and areas identified at the Colloquium for Information Systems Security Education (CISSE-2008), as well as a plan for the future development of a formalized research agenda for digital forensics.
- Garfinkel: Adopt standardized, modular approaches for data representation and digital forensic processing.

**CHALLENGES IN DIGITAL FORENSICS (Karie and Venter 2015)**

1. **Technical Challenges**
    i. Encryption
    ii. Vast Volumes of Data
    iii. Incompatibility among Heterogeneous Forensic Tools
    iv. Volatility of Digital Evidence
    v. Bandwidth Restrictions
    vi. Limited Lifespan of Digital Media
    vii. Sophistication of Digital Crimes
    viii. Emerging Technologies and Devices
    ix. Limited Window of Opportunity for Collection of Potential Digital Evidence
    x. Anti-Forensics
    xi. Acquisition of Information from Small-Scale Technological Devices
    xii. Emerging Cloud Computing or Cloud Forensic Challenges
2. **Legal Systems and/or Law Enforcement Challenges**
    i. Jurisdiction
    ii. Prosecuting Digital Crimes (Legal Process)
    iii. Admissibility of Digital Forensic Tools and Techniques
    iv. Insufficient Support for Legal Criminal or Civil Prosecution
    v. Ethical Issues
    vi. Privacy

3. **Personnel-Related Challenges**
    i. Lack of Qualified Digital Forensic Personnel (Training, Education, and Certification)
    ii. Semantic Disparities in Digital Forensics
    iii. Lack of Unified Formal Representation of Digital Forensic Domain Knowledge
    iv. Lack of Forensic Knowledge Reuse among Personnel
    v. Forensic Investigator Licensing Requirements
4. **Operational Challenges**
    i. Incident Detection, Response, and Prevention
    ii. Lack of Standardized Processes and Procedures
    iii. Significant Manual Intervention and Analysis
    iv. Digital Forensic Readiness Challenge in Organizations
    v. Trust of Audit Trails

Challenges can also be clustered and analyzed according to diverse categories for example standards, data acquisition, investigation, and reporting [28] (Figure 3.4).

Research identifies six main sources of errors in DF, as mentioned below [29]:

- Client errors.
- Wider investigative/forensic team errors.
- Practitioner errors.
- Tools/instruments errors.
- Method errors.
- Trace errors.

Machine learning contributes directly to trace errors, which occur at the investigation stage and focus on identifying and understanding digital traces such as data processing/examination, analysis, interpretation, and evaluation, and that may lead to "Missed data whereby erroneous traces may have a structure that ultimately means any tools and methods may fail to identify them (resulting in them being missed during an examination), misrepresentation: whereby traces may be parsed and displayed incorrectly impacting any subsequent review of them by practitioners, and general misinterpretation. Trace errors have the potential to compromise the value of digital investigations" [30].

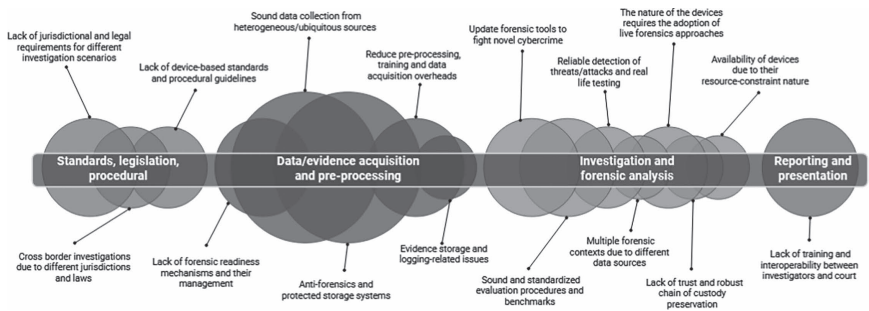

**FIGURE 3.4** Challenges in DF (Source: Casino et al. 2022).

In addition to the unintended consequence of bias and error, DF processes can be complicated further by explicit misrepresentation of the facts with the intention to deceive, such as fabrication, when evidence is made up, tampering when evidence is manipulated, or omission, where evidence is not taken into account. Either of these occurrences is common, difficult to identify, and compromise the validity of any outcome.

## 3.5 REASONING

One of the features of intelligence is, arguably, the ability to reason. Although reasoning has been a main branch of classical logic and a field of study in many disciplines since the beginning of human history, all the knowledge contained in the field of logic cannot explain the totality of a universe of discourse, with each approach having limitations that must be acknowledged [31].

The main types of reasoning are known as inductive, abductive, and deductive, whereby induction begins with data and produces concepts; in abduction [32], relationships among the concepts are inferred to develop interrelated hypotheses; and in deduction, data is gathered to fill in the gaps and produce hypotheses. Abduction is "process of forming an explanatory hypothesis. It is the only logical operation which introduces any new idea" [32].

Experts and analysts aiming to capture truth combine all three types of reasoning [29]. Thanks to advancements in computer science, the classification of reasoning types has been extended to include diagrammatic, abstract, spatial-temporal, fuzzy, and probabilistic reasoning. It is not in the scope of this paper to discuss logic in detail; however, being reasoning the backbone of all DF capabilities, one of its cornerstones is causal inference [33], described as the ability of an algorithm to establish causal relationships among factors being analyzed. In causal inference, confounds are fact*ors that influence both the dependent and independent variable*, causing spurious associations.

Confounding factors are inherent statistical errors caused by confounding data that naturally exist in any model. But they can also be fabricated and injected into the evidence deliberately to mislead inquiry. There is no known mechanism in DF to identify, represent, and address malicious confounding, and no reliable algorithms for its identification. Expert reasoning is characterized by heuristic informal judgmental rules which guide towards plausible paths to follow and away from implausible ones [34, 35]. In solving complex problems, investigators apply heuristic expert reasoning utilizing a mix of seemingly arbitrary logical composition, so-called fuzzy causal inference:

> *In the complex domain of causal inference, the challenge has always been to extract clear insights from the often vague and imprecise data coming from real-world scenarios. These frameworks have provided robust methodologies and points of views to find the cause-and-effect relationships from both observational and experimental*

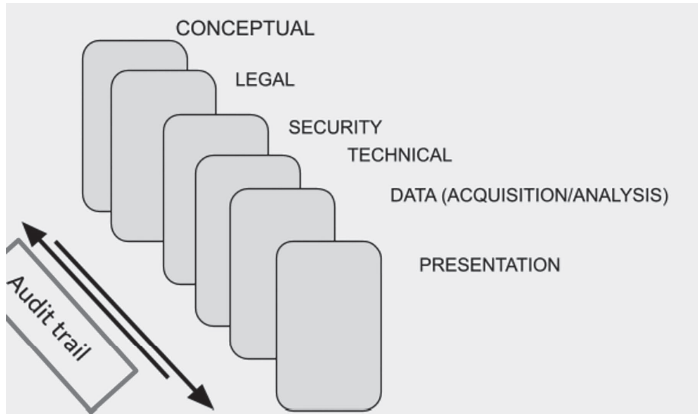

**FIGURE 3.5** DF Model layers (inspired by Renaud 2021).

> *data. However, they are in trouble for dealing with the inherent fuzziness of real-life data and the subjectivity of the human decision-making* [36].

Although still a relatively new approach, heuristic reasoning can, at least in part, be automated with fuzzy logic. The automation of hybrid logical schemas and fuzzy inference in DF using AI can deliver benefits and advance machine reasoning towards more sophisticated human-like intelligence, yet it requires careful and explicit representation.

When using machine learning to carry out complex deductive reasoning involving large digital datasets, a lot can go wrong. Structured analysis approaches can help guide the adherence of DF processes to the correct protocol, evaluate the correctness of the outcomes, and generate explicit, human-readable documentary evidence. A layered framework is an example of an explicit knowledge representation that can be used to mitigate bias and error, provided it can support an audit trail [37] across the layers to ensure process integrity (Figure 3.5).

## 3.6  TOWARDS OPEN STANDARDS

Open-source data and information have started to become accepted as admissible evidence in mainstream judicial proceedings where it has been referred to as digital open-source information (DOSI) [38]. Open standards can be crucial for ensuring that forensic practices are consistent, interoperable, and transparent across different tools and systems. They are intended to provide a common framework to facilitate the reliable exchange of data and results between various stakeholders. Yet, in reality, as of the time of writing, open standards for digital forensics are still being tentatively drafted. The main DF standards in use remain proprietary and as such require a license to be accessed and used (see Table 4.1). A notable exception is DF XML [39], which is currently maintained by National Institute of Standards and Technology (NIST)[2] and which is limited to representing metadata and provenance information of a subset of DF entities such as disk images; files; file system metadata; moves, adds, and changes (MAC) times; file hashes; sector hashes; transmission control protocol (TCP) flows; and hash sets.

DF XML does not provide a standardized representation of the state of the art in DF and AI at the time of writing.

**Main standards in DF:**

- ISO/IEC 27043:2015 – "Incident investigation principles and processes": outlines the principles and processes for conducting thorough and effective investigations of digital incidents.
- NIST Special Publication 800–101 – "Guidelines on Mobile Device Forensics": This publication provides guidelines for the forensic examination of mobile devices, including smartphones and tablets.
- NIST Special Publication 800–86 – "Guide to Integrating Forensic Techniques into Incident Response": This guide outlines how to integrate forensic techniques into incident response procedures to improve the effectiveness of investigations.
- NIST Special Publication 800–92 – "Guide to Computer Security Log Management": This publication provides guidance on the management and analysis of computer security logs, which are crucial for forensic investigations.
- ACPO (Association of Chief Police Officers) Guidelines – "Good Practice Guide for Computer-Based Evidence": These guidelines provide best practices for handling computer-based evidence to ensure that it remains intact and reliable.
- SANS Institute's **** "Forensic Readiness" : This guide emphasizes the importance of preparing for potential forensic investigations by implementing practices that facilitate the recovery and analysis of digital evidence.
- SWGDE (Scientific Working Group on Digital Evidence) Guidelines – "Digital Evidence Guidelines": These guidelines cover various aspects of digital evidence handling and analysis, promoting consistency and reliability in forensic practices.
- Forensic Science Regulator's Code of Practice and Conduct (UK) – "Code of Practice and Conduct for Digital Forensics": This code provides standards for digital forensic practices in the UK, ensuring that forensic work is conducted to a high standard.
- **ASTM E1188–11** – Standard Practice for Collection and Preservation of Information and Physical Items by a Technical Investigator.

There are innumerable known benefits that can be derived from the adoption of open technical standards; above all, increased transparency [40], interoperability [41], encouraging innovation [42], and reducing costs associated with proprietary technologies and vendor lock-in. Organizations can avoid costly licensing fees and choose from a variety of solutions that comply with the open standard, leading to more competitive pricing [43].

Machine readability can facilitate the automation of parsing documents by computers, but to evaluate and use technical and engineering artifacts, they must be human-readable. Readability metrics of technical documentation include reading speed, text comprehension [44], and the ability to interpret, comment on, and contribute to the technical document [45].

Human readability in AI can be achieved using natural language and visualizations as well as user-friendly interfaces to query and analyze the algorithms in use. Therefore, the case for human-readable DF artifacts through the adoption of human-friendly open standards supported by explicit logical schemas for DF is implicit and very strong. A human-readable model for DF that supports explicit knowledge representation [46] can serve as the outline of a model card and is proposed here as the basis for the development of an open standard in AI for DF.

Model cards (MC) are described as "short pieces of documentation that can be leveraged by ML applications to summarize and make transparent machine learning models, as well as executable programs to test/evaluate inputted data using a model" [47].

MCs are declarative frames [48]: explicit knowledge structures written in natural language that describe the knowledge (declarative) and processes (procedures) used by the AI. In declarative modeling, models declare a set of facts that are true about the model.

## 3.7 MODEL CARDS IN DIGITAL FORENSICS

Capturing and making explicit the many facets of DF in a single frame may not be feasible; however, a set of nested frames can be adopted – a series of nested model cards (DF MC9 and 1, for example).

An indicative example of how model cards can be designed to represent core elements of the DF process so that it can be standardized and made transparent is illustrated in Figures 3.6 and 3.7.

The DF MC 0 (Figure 3.6) includes the representation of data types, reasoning types, and sources of errors. Using a symbolic representation (in the form of natural language or pseudocode description):

**DF ModelCard_Elements (List)**
- MMCID – Identifier
- MCV – Version
- DF-MCO Owner
- DF-MC Use (standalone, integrated)
- DF-MC CS (Case statement)
- DF-MC H (Hypothesis)
- DF-MC C (Classification)
- DF-MC TR (Type of reasoning)
- DF-MC B (Bias)
- DF-MC CB (Cause of Bias)

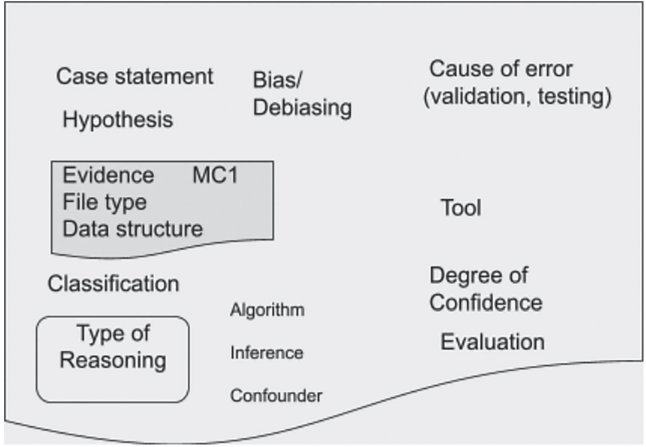

**FIGURE 3.6(a)**    DF MC 0 – Top Level.

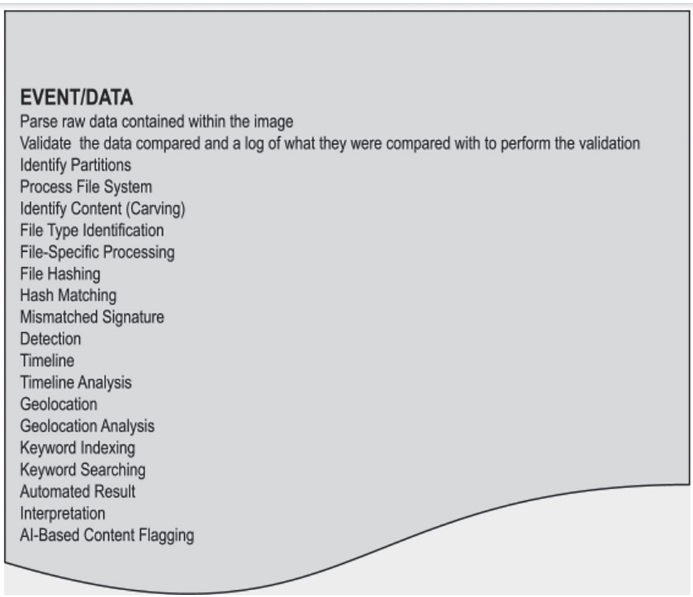

**FIGURE 3.6(b)**    DF MC 1  Includes a specification of data and processes associated with the data; DF MC 1 = DF Data types and analytical processes (based on Heargraves 2024).

- DF-MC E (Error)
- DF-MC CE (Cause of Error)
- DF-MC Ln (Layer n)
- Others

## 3.8  FUTURE WORK

DF as a discipline can be advanced by machine learning, provided the formalisms adopted adequately support the multidimensionality and transdisciplinarity of the domain by understanding the criticality of its systemic complexity.

AI must be designed to address and correct known flaws in DF, rather than reinforcing them.

How can machine learning be used in DF so that it is not used to support miscarriages of justice nor to distort truth or deviate from ethical and just conclusions and decision-making?

There is no simple answer. If machine learning is to improve DF processes and make them more transparent and accountable, and to mitigate human errors rather than automate their propagation, then the fairness of the line of inquiry must be unambiguously represented, made explicit, and easily auditable.

This chapter analyzes and summarizes the state of the art, paying particular attention to the current challenges in DF and AI, providing pointers to issues that limit the reliability of ML adoption. It contributes arguments and justification towards the development of human-readable open standards and provides the outline of model cards as a schema grounded in state-of-the-art research that captures the multiplicity of dimensions provided for further discussion, validation, and refinement. Future work includes formal evaluation and proposing the model card and elements as an extension of existing standards.

**About the Author:** Paola Di Maio is a systems analyst, independent researcher, and a research scholar at the Institute for Global Distributed Open Research and Education (IGDORE) and the Chair of the W3C AI KR Community Group.

## NOTES

1  https://www.nnlm.gov/guides/data-glossary/data-management-plan
2  https://www.loc.gov/preservation/digital/formats/fdd/fdd000611.shtml

## REFERENCES

[1] Casey, E. (2009). *Handbook of digital forensics and investigation*. Academic Press.
[2] Cranor, L. F. (2008). A framework for reasoning about the human in the loop. *UPSEC'08: Proceedings of the 1st conference on usability, psychology, and security* (pp. 1–15). https://dl.acm.org/doi/10.5555/1387649.1387650
[3] Garfinkel, S. (2010). Digital forensics research: The next 10 years. *Digital Invest*, 7, S64–73.
[4] Himmelsbach, T., Mou, Y., Decker, S., & Heinzl, A. (2023). Towards federated machine learning and distributed ledger technology-based data monetization. In *VLDB workshops*. ceur-ws.org
[5] Jones, G. M., & Winster, S. G. (2022). An insight into digital forensics: History, frameworks, types and tools. *Cyber Security and Digital Forensics,* 105–125.
[6] Ahmed, R., & Dharaskar, R. V. (2008). Mobile forensics: An overview, tools, future trends and challenges from law enforcement perspective." In *6th international conference on e-governance, ICEG, emerging technologies in e-government, m-government* (pp. 312–323). http://www.iceg.net/2008/. 17-19 April 2009 3rd IEEE/ACM
[7] Meghanathan, N., Allam, S. R., & Moore, L. A. (2010). Tools and techniques for network forensics. *arXiv preprint arXiv:1004.0570.*

[8] Murdoch, D., Xu, J., Liu, S., Shen, Y., & Yu, C. (2018). Big data forensics: Challenges and opportunities. *IEEE Cloud Computing*, *5*(1), 50–57.

[9] Noyes, J., & Keegan, M. (2019). Big data analytics in digital forensics. In A. Parisi & M. Piattini (Eds.), *Digital forensics and cyber crime* (pp. 289–305). Springer International Publishing.

[10] Himmelsbach, M., & Themis, M. (2020). The role of digital forensics in data science. In *International conference on cybercrime (ICCC)* (pp. 122–129). IEEE; Garfinkel, S. (2010). *Digital forensics science: The best practices illustrated*. Addison-Wesley Professional.

[11] Dunsin, D., Ghanem, M. C., Ouazzane, K., & Vassilev, V. (2024). A comprehensive analysis of the role of artificial intelligence and machine learning in modern digital forensics and incident response. *Forensic Science International: Digital Investigation*, *48*, 301675.

[12] Al-Jadir, I., Wong, K. W., Fung, C. C., & Xie, H. (2018). October. Enhancing digital forensic analysis using memetic algorithm feature selection method for document clustering. In *2018 IEEE international conference on systems, man, and cybernetics (SMC)* (pp. 3673–3678). IEEE.

[13] Dimitriadis, A., Lontzetidis, E., Kulvatunyou, B., Ivezic, N., Gritzalis, D., & Mavridis, I. (2022). Fronesis: Digital forensics-based early detection of ongoing cyber-attacks. *IEEE Access*, *11*, 728–743.

[14] Zhang, Q., Yang, L. T., Chen, Z., & Li, P. (2018). A survey on deep learning for big data. *Information Fusion*, *42*, 146–157.

[15] Faehndrich, J., Honekamp, W., Povalej, R., Rittelmeier, H., Berner, S., & Labudde, D. (2023). Digital forensics and strong AI: A structured literature review. *Forensic Science International: Digital Investigation*, *46*, 301617.

[16] Cole, S. A. (2014). *Forensic science and miscarriages of justice*. Springer Science+Business Media.

[17] Etter, B. (2013). The contribution of forensic science to miscarriage of justice cases. *Australian Journal of Forensic Sciences*, *45*(4), 368–380.

[18] Sunde, N., & Dror, I. E. (2019). Cognitive and human factors in digital forensics: Problems, challenges, and the way forward. *Digital Investigation*, *29*, 101–108.

[19] Dror, I. E. (2020). Cognitive and human factors in expert decision making: Six fallacies and the eight sources of bias. *Analytical Chemistry*, *92*(12), 7998–8004.

[20] Moser, S. (2013). Confirmation bias: The pitfall of forensic science. *Themis: Research Journal of Justice Studies and Forensic Science*, *1*, Article 7.

[21] Renaud, K., Bongiovanni, I., Wilford, S., & Irons, A. (2021). PRECEPT-4-Justice: A bias-neutralising framework for digital forensics investigations. *Science & Justice*, *61*(5), 477–492.

[22] Jinad, R., Gupta, K., Simsek, E., & Zhou, B. (2024). Bias and fairness in software and automation tools in digital forensics. *Journal of Surveillance, Security and Safety*, *5*, 19–35.

[23] Sunde, N., & Dror, I. E. (2019). Cognitive and human factors in digital forensics: Problems, challenges, and the way forward. *Digital Investigation*, *29*, 101–108.

[24] Karie, N. M., & Venter, H. S. (2015). Taxonomy of challenges for digital forensics. *Journal of Forensic Sciences*, *60*(4), 885–893.

[25] Mijwil, M., & Aljanabi, M. (2023). Towards artificial intelligence-based cybersecurity: The practices and ChatGPT generated ways to combat cybercrime. *Iraqi Journal For Computer Science and Mathematics*, *4*(1), 65–70.

[26] Dempsey, R. P., Brunet, J. R., & Dubljević, V. (2023). Exploring and understanding law enforcement's relationship with technology: A qualitative interview study of police officers in North Carolina. *Applied Sciences*, *13*(6), 3887.

[27] Swofford, H., & Champod, C. (2022). Probabilistic reporting and algorithms in forensic science: Stakeholder perspectives within the American criminal justice system. *Forensic Science International: Synergy*, *4*, 100220.

[28] Casino, F., Dasaklis, T. K., Spathoulas, G. P., Anagnostopoulos, M., Ghosal, A., Borocz, I., Solanas, A., Conti, M., & Patsakis, C. (2022). Research trends, challenges, and emerging topics in digital forensics: A review of reviews. *IEEE Access*, *10*, 25464–25493

[29] Åsvoll, H. (2013). Abduction, deduction and induction: Can these concepts be used for an understanding of methodological processes in interpretative case studies? *International Journal of Qualitative Studies in Education*, *27*(3), 289–307. https://doi.org/10.1080/09518398.2012.759296.

[30] Horsman, G. (2024). Sources of error in digital forensics. *Forensic Science International: Digital Investigation*, *48*, 301693.

[31] Yanofsky, N. S. (2016). *The outer limits of reason: What science, mathematics, and logic cannot tell us*. MIT Press.

[32] Peirce, C. S. (1992). *The essential Peirce, volume 2: Selected philosophical writings (1893–1913)* (Vol. 2). Indiana University Press.

[33] Yao, L., Chu, Z., Li, S., Li, Y., Gao, J., & Zhang, A. (2021). A survey on causal inference. *ACM Transactions on Knowledge Discovery from Data (TKDD)*, *15*(5), 1–46.e

[34] Polya, G. (1945). *How to solve it*. Princeton, NJ: Princeton University Press.

[35] Lenat, D. B. (1982). The nature of heuristics. *Artificial Intelligence*, *19*(2), 189–249.

[36] Saki, A., & Faghihi, U. (2024). Integrating fuzzy logic with causal inference: Enhancing the pearl and Neyman-Rubin methodologies. *arXiv preprint arXiv:2406.13731*.

[37] Surden, H. (2019). Machine learning and the law. *Harvard Journal on Legislation*, *56*(1), 25–65.

[38] Gillett, M., & Fan, W. (2023). Expert evidence and digital open source information: Bringing online evidence to the courtroom. *Journal of International Criminal Justice*, *21*(4), 661–693.

[39] Garfinkel, S. (2012). Digital forensics XML and the DFXML toolset. *Digital Investigation*, *8*(3–4), 161–174.

[40] Russell, A. L. (2014). *Open standards and the digital age*. Cambridge University Press.

[41] Berners-Lee, T. (2005). *Weaving the web: The original design and ultimate destiny of the world wide web by its inventor*. San Francisco: Harper.

[42] Mowery, D. C., & Rosenberg, N. (1991). *Technology and the pursuit of economic growth*. Cambridge University Press.

[43] David, P. A., & Foray, D. (2003). Economic fundamentals of the knowledge society. *Policy Futures in Education*, *1*(1), 20–49.

[44] Kriukova, A. (2022). *Measuring readability of technical texts*. Institute of Formal and Applied Linguistics (Master thesis).

[45] Oliveira, D., Bruno, R., Madeiral, F., & Castor, F. (2020, September). Evaluating code readability and legibility: An examination of human-centric studies. In *2020 IEEE international conference on software maintenance and evolution (ICSME)* (pp. 348–359). IEEE.

[46] Hargreaves, C., Nelson, A., & Casey, E. (2024). An abstract model for digital forensic analysis tools-A foundation for systematic error mitigation analysis. *Forensic Science International: Digital Investigation*, *48*,

[47] Mitchell, M., Wu, S., Zaldivar, A., Barnes, P., Vasserman, L., Hutchinson, B., Spitzer, E., Raji, I. D., & Gebru, T. (2019, January). Model cards for model reporting. In *Proceedings of the conference on fairness, accountability, and transparency* (pp. 220–229). ACM. Org.

[48] Winograd, T. (1975). Frame representations and the declarative/procedural controversy. In *Representation and understanding* (pp. 185–210). Morgan Kaufmann.